\documentclass[runningheads]{llncs}
\usepackage{todonotes}
\usepackage{graphicx}
\usepackage{amsmath}
\begin{document}
%
\title{Scalable Cloud-Native Pipeline for Efficient 3D Model Reconstruction from Monocular Smartphone Images}
%
\titlerunning{Scalable Cloud-Native 3D Model Reconstruction Pipeline from 2D Images}
%
\author{Aghilar Potito\inst{1} \and Anelli Vito Walter\inst{2} \and \\Trizio Michelantonio\inst{1} \and Di Noia Tommaso\inst{2}}
\authorrunning{P. Aghilar et al.}
%
\institute{Wideverse, Politecnico di Bari, 70125, Italy \\ 
\email{\{potito,mike\}@wideverse.com}\\
\url{https://www.wideverse.com/} \and
SisInfLab, Politecnico di Bari, 70125, Italy \\
\email{\{vitowalter.anelli,tommaso.dinoia\}@poliba.it}\\
\url{https://sisinflab.poliba.it/}}

%
\maketitle              
\begin{abstract}
In recent years, 3D models have gained popularity in various fields, including entertainment, manufacturing, and simulation. However, manually creating these models can be a time-consuming and resource-intensive process, making it impractical for large-scale industrial applications. To address this issue, researchers are exploiting Artificial Intelligence and Machine Learning algorithms to automatically generate 3D models effortlessly. In this paper, we present a novel cloud-native pipeline that can automatically reconstruct 3D models from monocular 2D images captured using a smartphone camera. Our goal is to provide an efficient and easily-adoptable solution that meets the Industry 4.0 standards for creating a Digital Twin model, which could enhance personnel expertise through accelerated training. We leverage machine learning models developed by NVIDIA Research Labs alongside a custom-designed pose recorder with a unique pose compensation component based on the ARCore framework by Google. Our solution produces a reusable 3D model, with embedded materials and textures, exportable and customizable in any external 3D modelling software or 3D engine. Furthermore, the whole workflow is implemented by adopting the microservices architecture standard, enabling each component of the pipeline to operate as a standalone replaceable module.

\end{abstract}
\setcounter{footnote}{0}
\section{Introduction}

In contemporary times, 3D models and complete 3D environments have become ubiquitous across different sectors, including art, entertainment, simulation, augmented reality, virtual reality, video games, 3D printing, marketing, TV and manufacturing. The attraction of having a digital version of any physical object as a 3D model lies in its versatility and adaptability for varied purposes. This digital replica, known as a \emph{Digital Twin (DT)}, is a virtual model that accurately reflects and maps physical goods in a digital space~\cite{F_Tao_et_al_2018}. DTs can be utilized to replicate physical objects in a virtual environment, thereby enabling the performance of specific tasks on the simulated model and observing their effects on the real-world counterpart. DTs have extensive applications in the manufacturing industry, including product design, process improvement, and optimization. Moreover, the integration of Industrial Augmented Reality (IAR) in Industry 4.0 can significantly enhance worker productivity and task effectiveness by providing real-time data and information, aiming to improve the overall operational efficiency. IAR is beneficial in manufacturing, where it assists workers in making informed decisions in realistic situations~\cite{moloney2006augmented}, streamlines engineering workflows throughout the design and manufacturing stages~\cite{schneider2017augmented}, and increases productivity by equipping workers with the necessary information to perform tasks more efficiently and safely. IAR is also effective in marketing and sales, where it can provide interactive information about products, dispel uncertainty, and influence client perceptions\cite{zhang2000commerce,hauswiesner2013virtual,wiwatwattana2014augmenting,el2016mobile}. Furthermore, IAR can facilitate training by offering detailed instructions and reducing the time required to train new personnel while minimizing their skill requirements\cite{hovrejvsi2015augmented}.

Over the years, modeling techniques have undergone significant evolution, leading to the development of more intuitive and less time-consuming tools for creating or generating 3D models. These models can be created from a set of primitive shapes, mathematical equations, or even a 2D image. The most commonly used techniques for creating 3D models are manual modeling, photogrammetry, and Light Detection and Ranging (LIDAR). Manual modeling, while effective, can be expensive in terms of time and resources since it involves a significant amount of manual labor and is unsuitable for large-scale applications. Alternatively, photogrammetry involves the use of photographs taken from different angles by a camera to make measurements. Finally, specialized hardware-based techniques such as LIDAR technology are also utilized.

In addition, the industrial research sector is actively exploring this research domain. For instance, NVIDIA is currently developing novel Artificial Intelligence and Machine Learning techniques and algorithms to enhance the quality of generated 3D models. Two of the approaches analyzed in this paper are based on recent research publications from 2022: \emph{Instant NeRF} - a set of instant neural graphics primitives for NeRF~\cite{mueller2022instant} - and \emph{nvdiffrec} - which leverages differential rendering and Deep Marching Tetrahedra (DMTet)~\cite{munkberg2021nvdiffrec}.

The aim of this paper is to present a distributed, cloud-native, and scalable pipeline capable of solving the 3D model reconstruction problem using a set of monocular two-dimensional images. The proposed pipeline is designed to reduce time and resources, providing a cost-effective solution for large-scale industrial applications by leveraging microservices architecture standards. Furthermore, the pipeline is enhanced by Augmented Reality (AR) capabilities to improve the data acquisition workflow.

The main contributions of this paper are:
\begin{itemize}
\item definition of a scalable cloud-native pipeline for the automatic generation of 3D models from monocular two-dimensional images with respect to the microservices architecture standard;
\item design and implementation of a custom pose recorder component based on ARCore to acquire both images of the object and poses of the camera.
\end{itemize}

\section{Background and Technology}

This section provides a detailed list of conventional and standard techniques alongside AI-based ones. It focuses on main drawbacks and how to overcome them.

\subsection{Standard and conventional techniques}

Manual modeling involves creating a 3D model using specialized software by an experienced 3D artist or modeler. This technique can be time-consuming and not suitable for large-scale applications due to the time involved for the design process for a single 3D model. The 3D artist is responsible for addressing various issues during the modeling process, such as mesh creation, material definition, texture generation, model rigging, environment, and lighting. Commonly used software for manual modeling includes techniques such as \emph{polygonal modeling}, \emph{surface modeling}, and \emph{digital sculpting}~\footnote{Wikipedia 3D modeling, \url{https://en.wikipedia.org/wiki/3D\_modeling,2022}}.

Photogrammetry is a technique for generating 3D models from two-dimensional images. It involves using a collection of photos taken from different angles with a standard 2D camera and extracting material properties using methods from optics and projective geometry. This technique is useful in achieving a realistic feeling during Physically-Based Rendering (PBR)~\footnote{Wikipedia Photogrammetry, \url{https://en.wikipedia.org/wiki/Photogrammetry}}.

Lastly, LIDAR is a remote sensing technology that uses pulsed laser light to measure variable distances from a source point to a hit point, thereby collecting data about the shape and elevation of the scanned object's surface. LIDAR is commonly used in 3D model reconstruction of real-world objects and is also known as a 3D laser scanner. The output of a LIDAR scan is a point cloud, which comprises a set of geo-located colored data points in a 3D space and provides additional information about the object's material properties~\footnote{Wikipedia LIDAR, \url{https://en.wikipedia.org/wiki/Lidar}}.

\subsection{AI-based techniques}

This paragraph discusses about how AI-based techniques can be used to overcome the aforementioned standard techniques' drawbacks. In particular, \emph{Instant NeRF} and \emph{nvdiffrec} from NVIDIA Research Labs~\cite{mueller2022instant,munkberg2021nvdiffrec}.

\subsubsection{Instant NeRF.}

It is a more advanced and efficient implementation of the NeRF technique, which enables the creation of 3D models from 2D images using neural networks and a multi-resolution hash encoding grid. The technique involves reconstructing a volumetric radiance-and-density field from 2D images and their corresponding camera poses, which can then be visualized through ray marching. The encoding phase is task-agnostic and only the hash table size is adjusted, which affects the trade-off between quality and performance. The multi-resolution structure enables the network to resolve hash collisions more effectively. The implementation heavily relies on parallelism, utilizing fully-fused CUDA kernels with \emph{FullyFusedMLP}\cite{tiny-cuda-nn,Müller2021}. If this is not available, the algorithm falls back to \emph{CutlassMLP} - CUDA Templates for Linear Algebra Subroutines\footnote{A. Kerr, D. Merrill, J. Demouth and J. Tran, “CUTLASS: Fast Linear Algebra in CUDA C++”, \url{https://developer.nvidia.com/blog/cutlass-linear-algebra-cuda/}, 2017}~\cite{Cutlass_Repo}, with a focus on minimizing unnecessary bandwidth and computational operations. The tests were conducted with a resolution of 1920×1080 on high-end hardware equipped with an NVIDIA RTX 3090 GPU with a 6MB L2 cache.

The primary limitation of this methodology is its dependence on the NeRF technique, which produces a point cloud as its output. Consequently, the authors had to devise a method to extract the mesh of the scene from the encoded data within the neural networks. To accomplish this, they employed the Marching Cubes (MC) algorithm, a mesh extraction technique that is dependent on a point cloud as its initial input. However, the resulting mesh presents surface irregularities in the form of various holes, lacks UV coordinates, and does not possess any materials. As a result, it is essentially an unusable gray mesh for any 3D modeling software.

\subsubsection{nvdiffrec.}

It is a tool that enables the creation of 3D models from 2D images. What sets nvdiffrec apart from Instant NeRF is its ability to reconstruct a 3D model surface, complete with texture and materials. The authors approached this task as an “inverse rendering” problem, using a 2D image loss function to optimize as many steps as possible jointly. The goal is to ensure that the reconstructed model's rendered images are of high quality compared to the input imagery. The approach used in nvdiffrec enables the learning of topology and vertex positions for a surface mesh without the need for any initial guesses about the 3D geometry. The tool's \emph{differentiable surface model} relies on a \emph{deformable tetrahedral mesh} that has been extended to support spatially varying materials and high dynamic range (HDR) environment lighting through a novel differentiable split sum approximation. The resulting 3D model can be deployed on any device capable of triangle rendering, including smartphones and web browsers, without the need for further conversion and can render at interactive rates~\cite{munkberg2021nvdiffrec}.

The paper tackles the challenge of 3D reconstruction from multi-view images of an object, with known camera poses and background segmentation masks, producing triangle meshes, spatially-varying materials (stored in 2D textures), and HDR environment probe lighting. Specifically, the authors adapt \emph{Deep Marching Tetrahedra (DMTet)} to work in the setting of 2D supervision and jointly optimize shape, materials, and lighting. Unlike Instant NeRF, the mesh in this approach is UV-mapped with customizable materials and multiple textures linked to it, allowing for the reuse of the mesh in any 3D engine, such as Blender~\footnote{Blender website, \url{https://www.blender.org/}}, Maya~\footnote{Maya website, \url{https://www.autodesk.it/products/maya/overview}}, 3DS Max~\footnote{3DS Max website, \url{https://www.autodesk.it/products/3ds-max/overview}}, and Unity~\footnote{Unity website, \url{https://unity.com/}}~\cite{munkberg2021nvdiffrec}.

\section{Proposed Pipeline}

Skilled service professionals are capable of maintaining and repairing complex machinery and industrial facilities. These professionals utilize their knowledge in industrial maintenance and assembly tasks by employing a combination of simulation, capturing techniques, multimodal interaction, and 3D-interactive graphics to achieve distributed training~\cite{WEBEL2013398}. The acquired competencies are then adapted to realistic training situations that are utilized in industrial training facilities. In~\cite{WEBEL2013398}, the authors refer to this as \emph{immersive training}, which involves \emph{“Real-time simulations of object behavior and multimodal interaction that support the development of complex training simulators that address cognitive skills [...] and sensorimotor skills.”}. Industrial Augmented Reality (IAR) is a combination of computer vision and computer graphics that utilizes camera-based interaction. IAR can be exploited to facilitate the data acquisition process for the proposed scalable cloud-native pipeline. A segment of the pipeline can be deployed within a Kubernetes cluster, where all cloud phases of the pipeline are dispatched as Jobs to worker nodes. Worker nodes require an NVIDIA GPU to handle the high-end capabilities needed for dataset preprocessing and reconstruction jobs. Therefore, the complete reconstruction pipeline consists of various phases that can be executed either on an embedded device or in the cloud, depending on the different resource requirements.

\subsection{Pipeline definition}

\begin{figure}
\includegraphics[width=\textwidth]{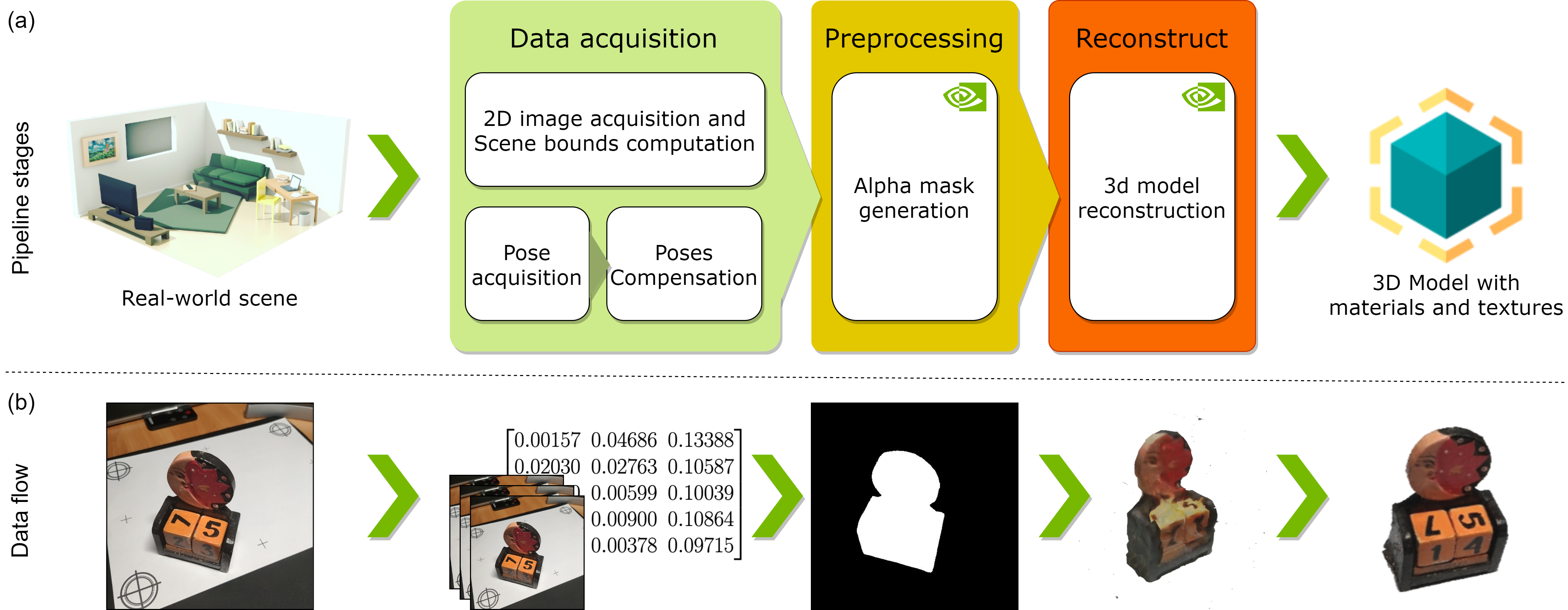}
\caption{A graphical representation of the proposed pipeline. In (a), the sequence of operations required to achieve the expected result are described. In (b), the data flow between the intermediate stages of the pipeline are illustrated.} \label{1_reconstruction_pipeline}
\end{figure}

We defined a reconstruction pipeline (Figure~\ref{1_reconstruction_pipeline}) by identifying a set of phases that are executed progressively, each performing specific operations on the dataset. The pipeline phases are described below:
\begin{itemize}
    \item \emph{dataset generation phase}, a custom written \emph{pose recorder} with a \emph{poses compensation algorithm} is implemented;
    \item \emph{dataset preprocessing phase}, the images and poses are preprocessed and the relative alpha masks are generated (silhouettes);
    \item \emph{reconstruction phase}, the 3D model is generated alongside a preview of the current pipeline status in order to provide feedback to the end user.
\end{itemize}
Upon completion of the entire process, the end user can interact with the generated 3D model and visualize it from different angles directly on his smartphone. We have adopted a storage solution that caches each intermediate output for the entire pipeline's flow in MinIO: an high-performance, S3 compatible, Kubernetes-native object storage solution~\footnote{MinIO website, \url{https://min.io/}}.

\subsection{Dataset generation phase}

The initial step in the reconstruction pipeline is the generation of the dataset, which comprises a collection of images and corresponding poses. These crucial components are obtained through a native Android application that implements the ARCore framework~\footnote{Google LLC, “ARCore SDK for Android”, \url{https://github.com/google-ar/arcore-android-sdk}}. The reconstruction module necessitates specific technical prerequisites for the input data, particularly:
\begin{itemize}
    \item a set of RGB images with a resolution of \( 512 \times 512 \) pixels;
    \item a set of alpha masks (silhouettes) with a resolution of \( 512 \times 512 \) pixels;
    \item a \emph{poses\_bounds.npy} file containing the view matrices of the camera for each image with the specific camera intrinsics.
\end{itemize}
Given a set of images of size \( N \), the \emph{poses\_bounds} file is a numpy~\footnote[5]{Numpy website - https://numpy.org/} array of shape \( (N, 17) \), in which \( N \) is the number of images and \( 17 \) is the number of total features for each image. The first 12 columns of each row are the \( 3 \times 4 \) \emph{view matrix} of the camera for the corresponding image, and the last 5 elements represent:
\begin{itemize}
    \item height of the image obtained from camera intrinsics;
    \item width of the image obtained from camera intrinsics;
    \item focal length of the camera obtained from camera intrinsics (we are assuming the focal lengths of both axes are the same);
    \item scene bounds obtained from depth map of the scene (the minimum and maximum distance from the camera).
\end{itemize}
It is imperative to maintain a coherent coordinate system throughout the entire process: both the ARCore framework and nvdiffrec adopt the same OpenGL right-handed system convention~\footnote{OpenGL Coordinate Systems, \url{https://learnopengl.com/Getting-started/Coordinate-Systems}}.

\subsubsection{Pose recorder.}

To record the poses during 2D image acquisition, a \emph{pose recorder component} is necessary, wherein each pose corresponds to a single image. We have implemented this workflow as a library in a native Android application, where the management of the anchor lifecycle is a critical aspect, particularly for the \emph{poses compensation algorithm}. Moreover, this library facilitates the selection of a camera with varying resolutions or frames per second (FPS) to initiate the recording process. In the subsequent \emph{preprocessing phase}, the images are resized to \( 512 \times 512 \) pixels to fulfill the input requirements of the machine learning model.

ARCore provides a view matrix of the device's pose in the world coordinate system, which is represented by a \( 4 \times 4 \) matrix. The rotation matrix is represented by the first \( 3 \times 3 \) submatrix, while the translation vector is represented by the last column. However, a \( 4 \times 4 \) matrix is not suitable for this particular problem, as a \( 3 \times 4 \) view matrix is required. To address this, the last row of the matrix is removed to obtain the desired \( 3 \times 4 \) matrix. The resulting matrix follows the \emph{column-major order} convention in which the matrix elements are ordered by column~\footnote{Wikipedia Row- and column-major order, \url{https://en.wikipedia.org/wiki/Row-\_and\_column-major\_order}}. To complete the transformation, a new column of shape \( 3 \times 1 \) that contains the height, width, and focal length of the device is concatenated with the matrix. The resulting matrix is:
\begin{equation*}
    \begin{bmatrix}
        r_{11} & r_{12} & r_{13} & t_x & h \\
        r_{21} & r_{22} & r_{23} & t_y & w \\
        r_{31} & r_{32} & r_{33} & t_z & f
    \end{bmatrix}
\end{equation*}
in which \( r_{ij} \) is the ij-element of the rotation view matrix, \( t_i \) is the i-element of the translation vector and \( h, w, f \) are the height, the width and the focal lenght respectively extracted from the camera instrinsics.
After a matrix flattening operation~\footnote{B. Mildenhall, “Test with known camera pose”, \url{https://github.com/Fyusion/LLFF/issues/10\#issuecomment-514406658}} and a subsequent concatenation, we obtain the final data-flattened view matrix. Thus, the following reshaped data entry can be generated for each frame:
\[ r_{11} \; r_{12} \; r_{13} \; t_x \; h \; r_{21} \; r_{22} \; r_{23} \; t_y \; w \; r_{31} \; r_{32} \; r_{33} \; t_z \; f \; m \; M \]
in which \( m \) and \( M \), are respectively the minimum and the maximum scene bounds computed from the depth map in meters. During the recording, a \emph{compensation matrix} is applied in real-time to compensate camera pose jumps. An additional rotational fix is applied to the \( 3 \times 3 \) rotation submatrix of the camera: it consists in a swap of the first and the second column and a sign inversion of the new first column (see footnote 13):
\begin{equation*}
    \begin{bmatrix}
        \textcolor{red}{r_{11}} & \textcolor{blue}{r_{12}} & r_{13} & t_x & h \\
        \textcolor{red}{r_{21}} & \textcolor{blue}{r_{22}} & r_{23} & t_y & w \\
        \textcolor{red}{r_{31}} & \textcolor{blue}{r_{32}} & r_{33} & t_z & f
    \end{bmatrix}
    \Rightarrow
    \begin{bmatrix}
        \textcolor{blue}{r_{12}} & \textcolor{red}{r_{11}} & r_{13} & t_x & h \\
        \textcolor{blue}{r_{22}} & \textcolor{red}{r_{21}} & r_{23} & t_y & w \\
        \textcolor{blue}{r_{32}} & \textcolor{red}{r_{31}} & r_{33} & t_z & f
    \end{bmatrix}
    \Rightarrow
    \begin{bmatrix}
        \textcolor{red}{-}r_{12} & r_{11} & r_{13} & t_x & h \\
        \textcolor{red}{-}r_{22} & r_{21} & r_{23} & t_y & w \\
        \textcolor{red}{-}r_{32} & r_{31} & r_{33} & t_z & f
    \end{bmatrix}
\end{equation*}
At the end of recording, different tasks are performed to generate the dataset:
\begin{itemize}
    \item the compensated \emph{poses\_bounds} file is saved in the device's local storage;
    \item the \emph{compensation matrix} is saved in the device's local storage;
    \item the images are cropped with an aspect ratio of 1:1 and saved in the device's local storage;
    \item the whole dataset is compressed and saved in the device's local storage.
    \item the compressed dataset is uploaded to the S3 bucket.
\end{itemize}

\subsubsection{Sensor drifting problem.}

During some tests, a jagged surface is observed in the reconstructed model, which indicates a misalignment of poses with the acquired images caused by a \emph{sensor drifting} problem. This problem generates an inconsistent dataset, posing a significant challenge for subsequent analysis. To address this issue, a comparison is performed between the generated \emph{poses\_bounds} file and the COLMAP generated one. COLMAP is an open-source software which implements \emph{Structure-from-motion (SfM)} and \emph{Multi-View Stereo (MVS) techniques}~\cite{schoenberger2016sfm,schoenberger2016mvs}. This comparison highlights the misalignment issue, leading to a solution to reach a more accurate and reliable dataset.

\begin{figure}
\includegraphics[width=\textwidth]{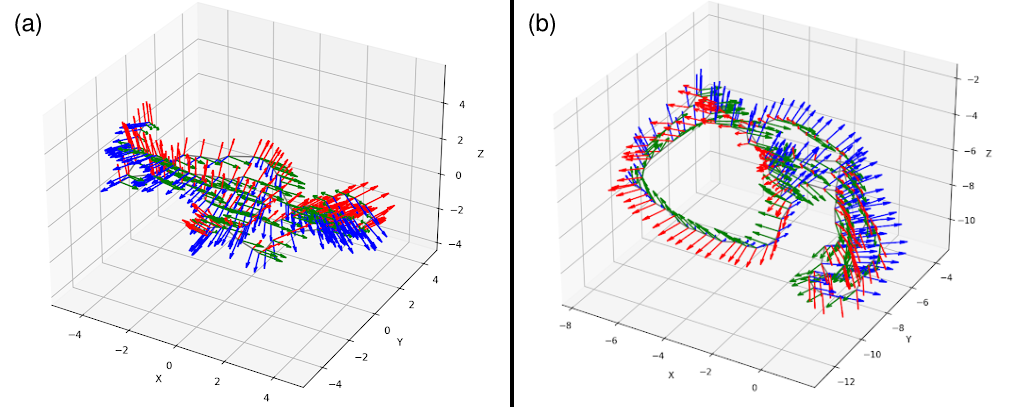}
\caption{Comparison of our solution's extracted poses (a) with COLMAP's (b). COLMAP lacks of real-world reference during the pose extraction phase resulting in a non-overlapped set of poses between (a) and (b).} \label{2_pose_vectors}
\end{figure}

The primary distinction between the two datasets stems from their distinct coordinate systems, which is due to the absence of a real-world reference in COLMAP (Figure~\ref{2_pose_vectors}). In order to reconcile all the data points, a series of transformations are implemented, wherein the entirety of the points are treated as a single rigid body. To achieve this, both rigid bodies are brought to a common origin, and a \emph{transformation matrix} is computed, which transforms three vectors from the COLMAP dataset to the ARCore one. The application of this transformation matrix affects both the rotation and scale, ultimately resulting in the overlapping of the two datasets. The computed difference between the two datasets yields a \emph{difference matrix}, which highlights their deviation.

\subsubsection{Pose compensation algorithm.}

\begin{figure}
\includegraphics[width=\textwidth]{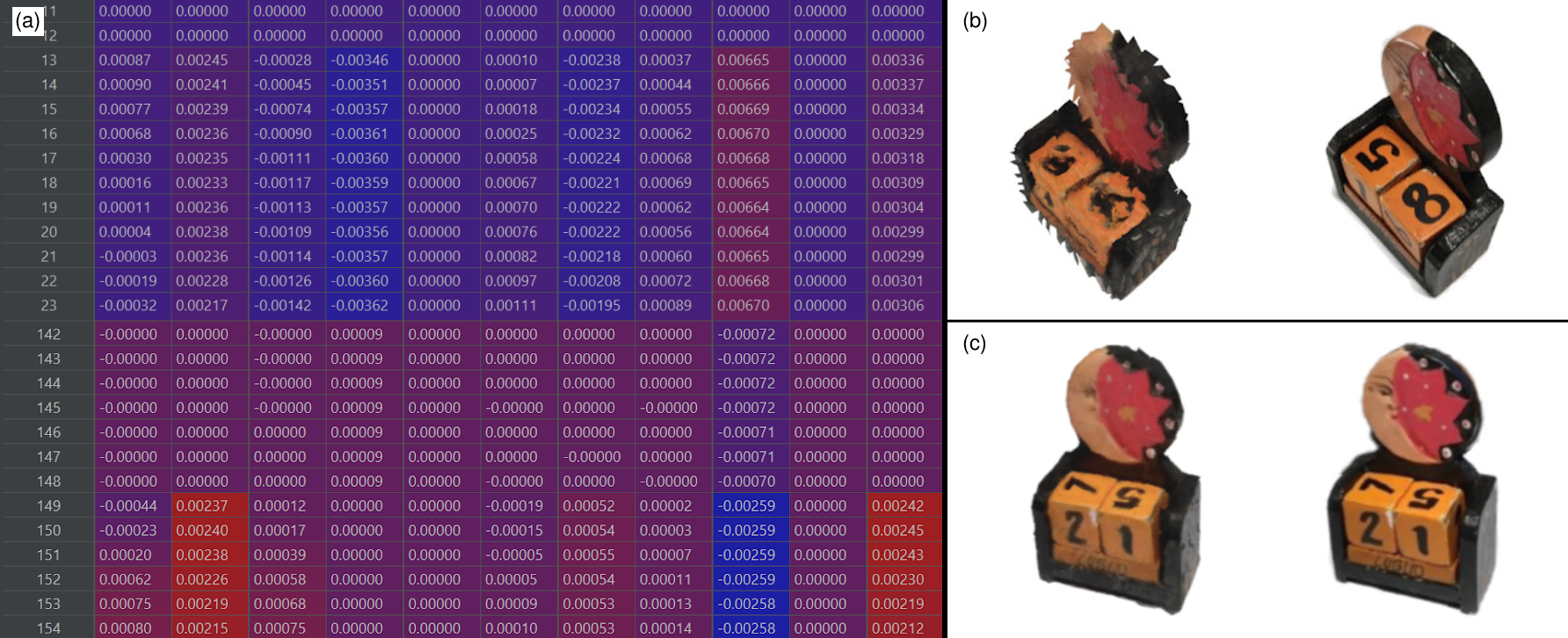}
\caption{In (a), a partial view of the compensation matrix generated at run-time is illustrated. In (b) and (c), the difference during reconstruction with the implementation of the compensation matrix is presented: in both cases the reference image is placed side by side to highlight the differences. } \label{3_compensation_matrix_reconstruction_before_and_after}
\end{figure}

The system relies on a self-made anchor management system to detect real-time variations of positions or rotations of ARCore Anchors while scanning. This avoids trajectory discontinuity by computing and applying a \emph{compensation matrix} to the camera view matrix. The Poses Compensation Algorithm comprises the following components:
\begin{itemize}
    \item \emph{Anchors}, objects placed in the scene, provided by ARCore;
    \item \emph{Delta position} from initial pose for each anchor frame by frame;
    \item \emph{Delta rotation} from initial pose for each anchor frame by frame;
    \item \emph{Quaternion} products to compute the rotation matrix.
\end{itemize}
Given a quaternion \( q = a + bi + cj + dk \) defined by the following coefficients \( <a, b, c, d> \) and the following imaginary components \( (i, j, k) \),  each delta quaternion can be computed as follows:
\[ q_{delta} = q_{target} q_{current}^{-1} \]
in which \( q_{delta} \) is the delta quanternion to compute, \( q_{target} \) represents the target rotation we want to reach and \( q_{current}^{-1} \) represents the inverse of the current rotation. Therefore, because \( q_{current}^{-1} \) is the conjugate of quaternion \( q_{current} \), it can be computed by an inversion of the imaginary components of the quaternion:
\[ conj(a + bi + cj + dk) = a - bi - cj - dk \]
Moreover, given two quaternions \( q \) and \( r \) having the form:
\[ q = q_0 + q_1i + q_2j + q_3k \quad \quad \quad r = r_0 + r_1i + r_2j + r_3k \]
From ~\cite{stevens_lewis_johnson}, the product of two quaternions is a quaternion having the form:
\[ n = q \times r = n_0 + n_1i + n_2j + n_3k  \]
where:
\[ n_0 = (r_0q_0 - r_1q_1 - r_2q_2 - r_3q_3) \quad \quad \quad n_1 = (r_0q_1 + r_1q_0 - r_2q_3 + r_3q_2) \]
\[ n_2 = (r_0q_2 + r_1q_3 + r_2q_0 - r_3q_1) \quad \quad \quad n_3 = (r_0q_3 - r_1q_2 + r_2q_1 + r_3q_0) \]
\\Specifically, the algorithm is composed of three main steps:
\begin{itemize}
    \item compute the \emph{delta mean pose} starting from the delta pose of each valid anchor (tracked from SDK): this indicates, on average, how much each anchor has moved from the initial pose. More anchors are placed in the scene, more accurate is the estimation;
    \item combine the current camera pose with the delta mean pose exploiting the \emph{pose composition} method;
    \item convert the new pose to a \( 3 \times 4 \) matrix and apply the rotational fix.
\end{itemize}
The values of the compensation matrix change frame by frame resulting in a full matrix of shape \( N \times 17 \) (Figure~\ref{3_compensation_matrix_reconstruction_before_and_after}).

\subsection{Dataset preprocessing phase}

In this phase, the images are resized to a resolution of \( 512 \times 512 \) pixels before starting the alpha masks generation subtask.

\subsubsection{Alpha masks generation.}

We adopted a machine learning model to extract the alpha mask starting from the RGB images, employing \emph{CarveKit}: an automated and high-quality framework for background removal in images using neural networks~\footnote{N. Selin, “CarveKit”, \url{https://github.com/OPHoperHPO/image-background-remove-tool}}. To optimize performance, the framework is executed on the GPU. Following this step, we refine the silhouettes by applying a threshold to eliminate any ambiguous regions and enhance the edges of the 3D model during the reconstruction phase. Ultimately, we obtain a set of sharpened alpha masks that are integrated into the initial dataset.

\subsection{Reconstruction phase}

This phase adopts nvdiffrec tool to reconstruct the 3D model. The input parameters required for this task are:
\begin{itemize}
    \item a collection of RGB images in PNG format;
    \item a collection of alpha mask images in PNG format (silhouette);
    \item a set of camera poses serialized in the \emph{poses\_bounds} numpy matrix file.
\end{itemize}
Finally, upon successful reconstruction, the tool provides as artifacts:
\begin{itemize}
    \item \emph{mesh.obj} containing the reconstructed mesh, UV mapped;
    \item \emph{mesh.mtl} containing the material properties;
    \item \emph{texture\_kd.png} file containing the diffuse texture;
    \item \emph{texture\_ks.png} containing the ORM map (-, roughness, metalness);
    \item \emph{texture\_n.png} containing the normal map.
\end{itemize}

\subsection{Architecture}

We have designed and implemented the entire pipeline utilizing microservices architecture standards, which has been specifically tailored for deployment on a Kubernetes cluster. In the upcoming sections, we will provide an in-depth description of the microservices involved in the process, as well as the cloud infrastructure adopted for this purpose.

\subsubsection{Microservices.}

\begin{figure}
\includegraphics[width=\textwidth]{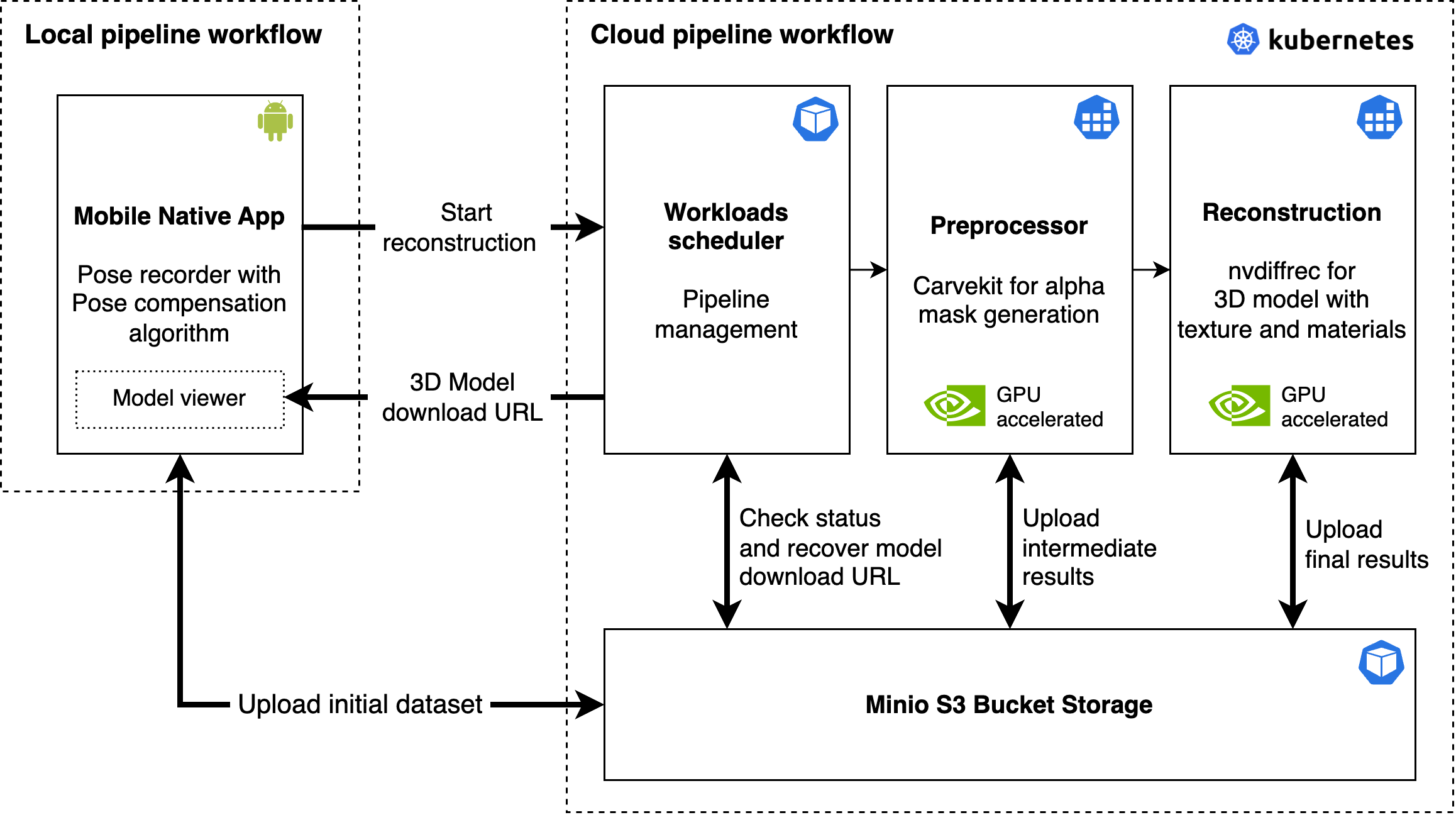}
\caption{Pipeline architecture with the Workloads scheduler, Preprocessor and Reconstruction microservices. The pipeline workflow is partitioned between local and cloud execution. All the stages communicate with the S3 storage layer to cache intermediate outputs and final 3D reconstructed model.} \label{7_architecture}
\end{figure}

The microservices compose the fundamental constituents of the pipeline. Each microservice, implemented as a Docker image, is purposefully crafted to accomplish a specific task. Specifically, the microservices that have been identified are the \emph{Preprocessor} microservice, the \emph{Reconstruction} microservice, and the \emph{Workloads scheduler} microservice (refer to Figure~\ref{7_architecture}).

The \emph{Preprocessor} microservice is dedicated to dataset preprocessing. Its container is based on the \emph{NVIDIA CUDA runtime environment} image, with all the necessary dependencies installed to ensure CarveKit to operate properly. The \emph{Reconstruction} microservice heavily relies on the nvdiffrec repository. This Docker image is built utilizing specific configurations, dependencies, and environmental variables outlined in the official documentation. The only modifications made are the addition of customized domain-specific code to enable preview image generation, dataset management, and 3D model uploading. Finally, the \emph{Workloads scheduler} microservice, on the other hand, is responsible for job scheduling within the Kubernetes cluster. It operates as a backend service API that oversees the entire pipeline lifecycle for each reconstruction request.

\subsubsection{Cloud infrastructure.}

The cloud infrastructure consists of a Kubernetes cluster deployed on bare metal, with accelerated machines designated as worker nodes. To support resource-intensive tasks such as dataset preprocessing and 3D model reconstruction, the worker nodes are equipped with an \emph{NVIDIA Quadro M4000 GPU}. However, due to certain specifications associated with the cloud nodes themselves, a \emph{Systemd-enabled Kubernetes worker CUDA-accelerated base image} was crafted leveraging a docker-in-docker execution~\footnote{S. Rana, “Docker and systemd”, \url{https://medium.com/swlh/docker-and-systemd-381dfd7e4628}}~\footnote{C. Zauner, “Running systemd inside a docker container”, \url{https://zauner.nllk.net/post/0038-running-systemd-inside-a-docker-container/}}.

\section{Evaluation}

This section provides an account of the performance and outcomes of the proposed solution. Furthermore, difficulties encountered during the study and prospects for enhancements are presented. Both qualitative and quantitative evaluations are included.

\subsection{Qualitative evaluation}

\begin{figure}[!ht]
\includegraphics[width=\textwidth]{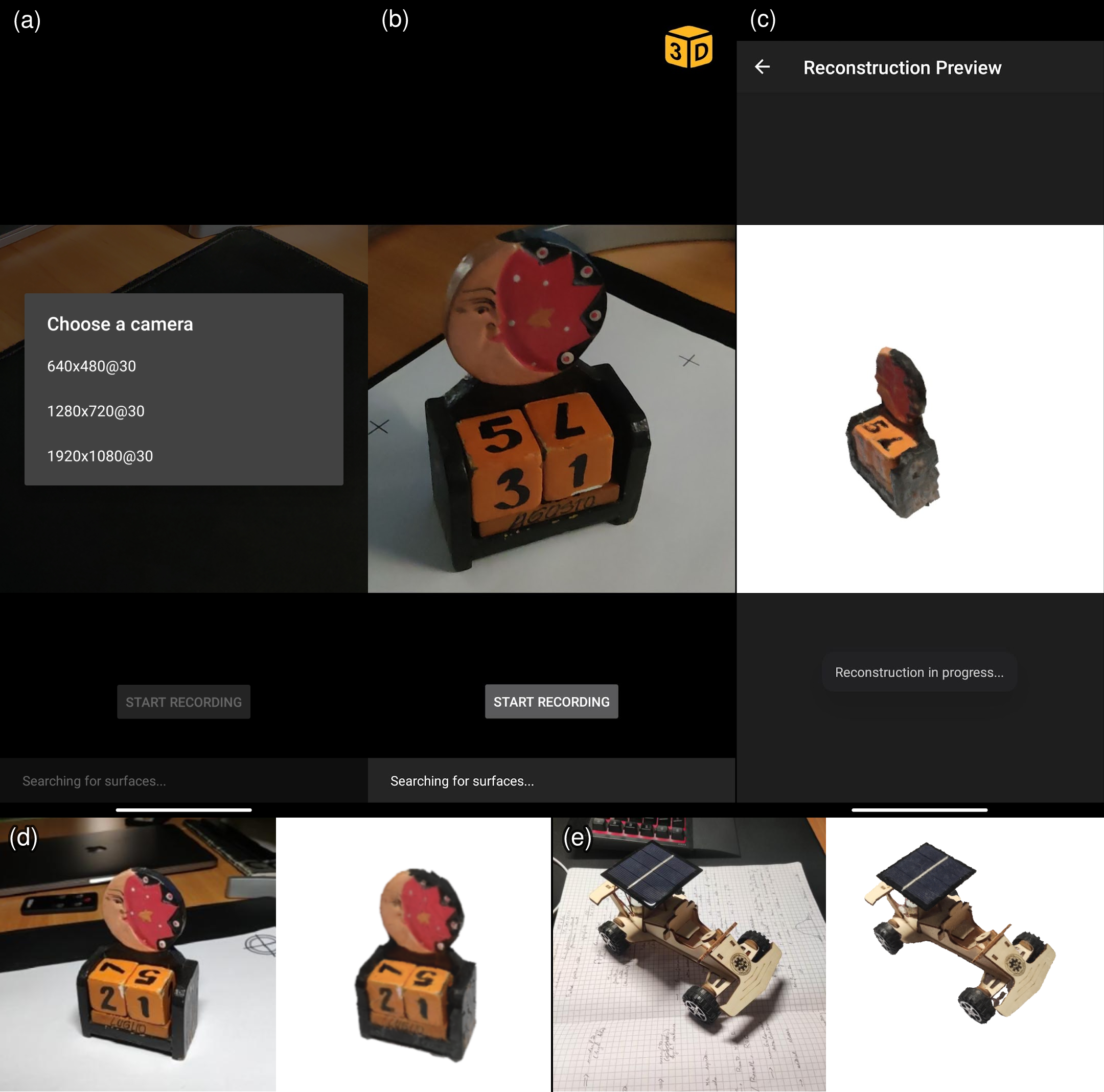}
\caption{Android application during camera selection (a), data acquisition (b) and reconstruction (c) phases. The whole pipeline workflow is transparent to the end user who is notified about the status through proper feedback on the User Interface (UI). In (d) and (e) two reconstruction attempts are depicted with their respective reference images.} \label{6_reconstruction_preview}
\end{figure}

The qualitative evaluation is performed considering user experience in mobile app utilization, alpha masks generation quality, and the real-look feeling of the generated 3D models (Figure~\ref{6_reconstruction_preview}).

The significance of user experience cannot be understated, especially when it comes to addressing challenges within specific industrial environments. A smartphone user is empowered to scan a variety of equipment, but it is crucial for them to be at ease with the requisite preparatory steps before initiating the model scanning process. In particular, the user must establish anchors by tapping on the screen next to distinguishable reference points. Failing to execute this preparatory process correctly may result in 3D models that are imprecise and of subpar quality. Moreover, it is imperative that the user receives updates on the progress of the reconstruction pipeline, along with a preview of the current model. In order to facilitate such feedback, the mobile application incorporates three status indicators.

The generation of alpha masks has a significant impact on the quality of the dataset, due to the silhouette extraction process involved. This procedure relies entirely on machine learning techniques, which are susceptible to errors such as inaccurate segmentation layers. As a result, it is imperative that the masks produced should be carefully examined by the user before commencing the reconstruction phase. This intermediate step, enables the erroneous alpha masks to be discarded from the dataset.

The two aforementioned steps have a significant impact on the overall quality of the dataset. As they are entirely reliant on machine learning techniques, errors may arise due to the lack of operator feedback. This can ultimately result in a reduced level of realism in the 3D models generated.

\subsection{Performance evaluation}

\begin{figure}
\includegraphics[width=\textwidth]{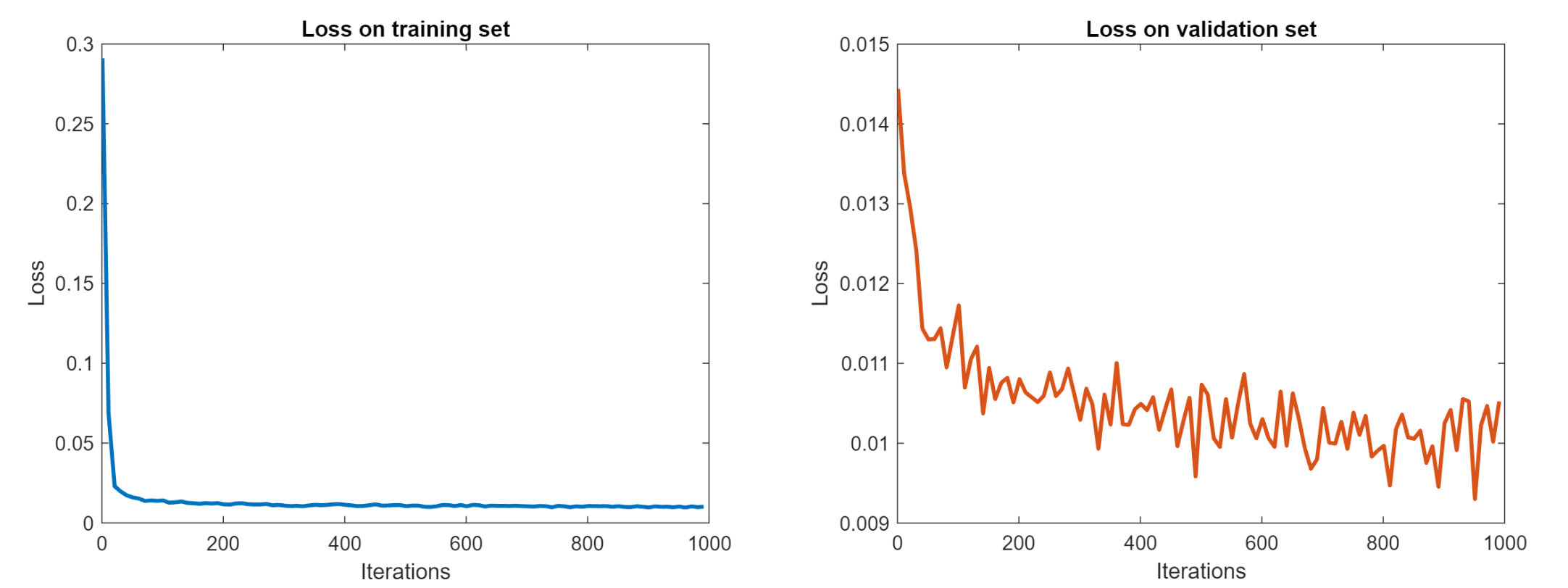}
\caption{The presented plots depict the loss values on the training and validation sets of the 3D reconstructed model performed using nvdiffrec. Both metrics were extracted from the aforementioned tool, serving as an evaluation of the model's performance.} \label{5_losses}
\end{figure}

The performance evaluation of the system takes into account the latency of the pipeline, from the scanning phase to the interaction phase. This latency can be computed using the following formula:
\[ T_{latency} = T_{scan} + 2T_{upload} + 2T_{signal} + T_{preprocessing} + T_{reconstruction} + T_{download} \]
Here, \( T_{signal} \) denotes the time required for signals to propagate within the infrastructure, while \( T_{upload} \) and \( T_{download} \) represent the time taken to upload and download assets from the S3 bucket, respectively. Since \( T_{scan} \), \( T_{preprocessing} \), and \( T_{reconstruction} \) take significantly longer than the other steps, we can simplify the formula as follows:
\[ T_{latency} = T_{scan} + T_{preprocessing} + T_{reconstruction} \]
In the conducted tests, network latency, denoted by \( T_{scan} \), was found to be approximately 120 seconds, while preprocessing time took roughly 30 seconds. The reconstruction process required approximately 2 hours and 30 minutes. Additionally, losses on both training and validation sets were taken into consideration. The calculation of image space loss was performed using nvdiffrast~\cite{Laine2020diffrast}, which assesses the difference between the rendered image and the reference one. As shown in Figure~\ref{5_losses}, the employment of nvdiffrec resulted in a loss of 0.010540 on the training set and 0.010293 on the validation set.

\section{Conclusion and future work}

This study presents an innovative cloud-native scalable pipeline for reconstructing 3D models of real-world objects, with the aim of producing 3D models for Digital Twins. This approach offers various advantages related to Industry 4.0, including a faster personnel training process. The proposed solution employs both low-end hardware, such as 2D cameras overlaid by Google's ARCore framework, and high-end cloud worker nodes for the segmentation and reconstruction tasks. Specifically, a machine learning model is adopted to segment the dataset. Once the alpha masks are generated, nvdiffrec tool by NVIDIA is exploited to perform the effective 3D model reconstruction. The resulting model can be downloaded and interactively viewed on a smartphone. The entire pipeline complies to microservices architecture standards, enabling scalability in large-scale production environments. Although the proposed solution has achieved the expected outcomes, the modular design allows for potential future improvements, including:
\begin{itemize}
    \item adoption of a better reconstruction machine learning model to produce smoother edges and better reconstructed models~\cite{Vicini2022sdf};
    \item replacement or improvement of the machine learning model used to generate the alpha masks;
    \item implementation of more layers to decompose the 3D model into its constituent parts, enabling a more exhaustive experience.
\end{itemize}

%
%
%
%

%
%
%
\end{document}